\documentclass[journal]{IEEEtran}
\usepackage{graphicx}
\usepackage{amsfonts}

\usepackage[english]{babel}
\usepackage{amsmath, amssymb}
\usepackage{float}
\usepackage{makecell}
\usepackage{multirow}
\usepackage{cite}
\usepackage{amsmath,amssymb,amsfonts}
\usepackage{algorithmic}
\usepackage{graphicx}
\usepackage{textcomp}
\usepackage{xcolor}

\usepackage{subfigure}
\usepackage{epstopdf}

\usepackage{graphicx}
\usepackage{comment}
\usepackage{amsmath,amssymb} 
\usepackage{color}
\usepackage{enumerate}

\usepackage{booktabs}

\ifCLASSINFOpdf
\else
\fi
\begin{document}
%
\title{
NccFlow: Unsupervised Learning of Optical Flow With Non-occlusion from Geometry
}

\author{Guangming~Wang,
       Shuaiqi Ren, and Hesheng Wang
        
\thanks{*This work was supported in part by the Natural Science Foundation of China under Grant U1613218, Grant U1913204, and Grant 62073222; in part by the Shanghai Municipal Education Commission and Shanghai Education Development Foundation through the Shu Guang Project under Grant 19SG08; and in part by grants from the NVIDIA Corporation. Corresponding author: Hesheng Wang.}
\thanks{Guangming Wang, Shuaiqi Ren, and Hesheng Wang are with the Key Laboratory of System Control and Information Processing of Ministry of Education, Department of Automation, Institute of Medical Robotics, Shanghai Jiao Tong University, Shanghai 200240, China, and also with the Key Laboratory of Marine Intelligent Equipment and System of Ministry of Education, Shanghai Engineering Research Center of Intelligent Control and Management, Shanghai Jiao Tong University, Shanghai 200240, China (e-mail: wanghesheng@sjtu.edu.cn).

}

}

%
%

\markboth{Journal of \LaTeX\ Class Files,~Vol.~14, No.~8, August~2015}%
{Shell \MakeLowercase{\textit{et al.}}: Bare Demo of IEEEtran.cls for IEEE Journals}
%



\maketitle


\begin{abstract}
Optical flow estimation is a fundamental problem of computer vision and has many applications in the fields of robot learning and autonomous driving. This paper reveals novel geometric laws of optical flow based on the insight and detailed definition of non-occlusion. Then, two novel loss functions are proposed for the unsupervised learning of optical flow based on the geometric laws of non-occlusion. Specifically, after the occlusion part of the images are masked, the flowing process of pixels is carefully considered and geometric constraints are conducted based on the geometric laws of optical flow. First, neighboring pixels in the first frame will not intersect during the pixel displacement to the second frame. Secondly, when the cluster containing adjacent four pixels in the first frame moves to the second frame, no other pixels will flow into the quadrilateral formed by them. According to the two geometrical constraints, the optical flow non-intersection loss and the optical flow non-blocking loss in the non-occlusion regions are proposed. Two loss functions punish the irregular and inexact optical flows in the non-occlusion regions. The experiments on datasets demonstrated that the proposed unsupervised losses of optical flow based on the geometric laws in non-occlusion regions make the estimated optical flow more refined in detail, and improve the performance of unsupervised learning of optical flow. In addition, the experiments training on synthetic data and evaluating on real data show that the generalization ability of optical flow network is improved by our proposed unsupervised approach.

\end{abstract}

\IEEEpeerreviewmaketitle

\section{Introduction}

\IEEEPARstart{O}{ptical} flow represents the 2D motion and correspondence relationship between two images at the pixel level, which is a fundamental problem in the field of computer vision. Optical flow has lots of applications in autonomous driving, such as  visual odometry \cite{min2020voldor}, target tracking \cite{ke2018real}, moving object detection, and mapping \cite{menze2018object,jiang2021moving}. In addition, the optical flow can be used to analyze the motion attributes of pedestrians and vehicles, so as to realize the dynamic understanding of scenes and decision-making. With the development of deep learning, good performance of optical flow estimation has been achieved by training on synthetic data \cite{dosovitskiy2015flownet,ilg2017flownet}. However, the gap between real data and synthetic data makes the supervised models on synthetic data have limited performance in real data. This spawned a large number of unsupervised studies of optical flow to make the trained optical flow network without the
gap when applied in real applications \cite{jason2016back,ren2017unsupervised,wang2018occlusion,meister2018unflow,alletto2018self,janai2018unsupervised,liu2019ddflow,liu2019selflow,zou2018df,ranjan2019competitive,wang2020unsupervised,jonschkowski2020matters}. Besides, the unsupervised method can utilize a large number of videos on the internet. The common basic idea behind the unsupervised learning of optical flow is based on the consistency between the target image and the reconstructed image. The reconstructed image is obtained by warping the source image utilizing the estimated optical flow field by the neural network model. Then, the neural network model is trained and updated to minimize the difference between the target image and the reconstructed image. This consistency assumption is not satisfied in the occlusion regions of images, so there are many previous works exploiting a lot of how to mask the occlusion regions \cite{wang2018occlusion,zou2018df,janai2018unsupervised,ranjan2019competitive,wang2020unsupervised,jonschkowski2020matters}. 

However, there are seldom studies on the constraints of optical flow on non-occlusion regions. The smoothness loss is for all optical flow in an image. The reconstruction loss utilizes the luminosity constraint, not considering the geometry of the optical flow. 
In this paper, it is found that there are also some geometric laws for the optical flow in the non-occlusion regions. To our best knowledge, this paper is the first to study the non-occlusion constraints of unsupervised optical flow learning. In this work, we reveal new geometric laws of the optical flow in non-occlusion regions and design two new unsupervised losses for the unsupervised learning of optical flow. Our contributions are as follows:

\begin{itemize}
	\item By carefully analyzing the motion of each pixel in real 3D space and 2D projected image, non-occlusion is defined in the 2D image in detail. New geometric laws of optical flow in the non-occlusion regions are revealed. 
	\item Based on the insight into the geometric laws of optical flow in the non-occlusion regions, two novel loss functions, the optical flow non-intersection loss and the optical flow non-blocking loss, are proposed for the unsupervised learning of optical flow. The non-intersection loss defines that optical flows should not cross each other in non-occlusion regions. The non-blocking loss defines that a pixel should not be surrounded by other nearby pixels during the pixel motion between adjacent frames.
	\item We integrate our proposed unsupervised losses into a unified framework of unsupervised optical flow and the experiments demonstrated the effectiveness of our proposed losses. The experiments on real dataset, KITTI 2015 dataset \cite{geiger2012we,menze2015joint}, show the good generalization ability of our model.
\end{itemize}

The rest of this paper is organized as follows. Section II summarizes the related works. Section III analysis the optical flow of each pixel from 3D to 2D and reveals new geometric laws of optical flow in the non-occlusion regions. The architecture of our unsupervised system and two novel loss functions based on the geometric laws are introduced in Section IV. The experiments details and results are in Section V. And Section VI concludes this paper.

\begin{figure*}[t]
	\centering
	\resizebox{1.00\textwidth}{!}
	{
		\includegraphics[scale=1.00]{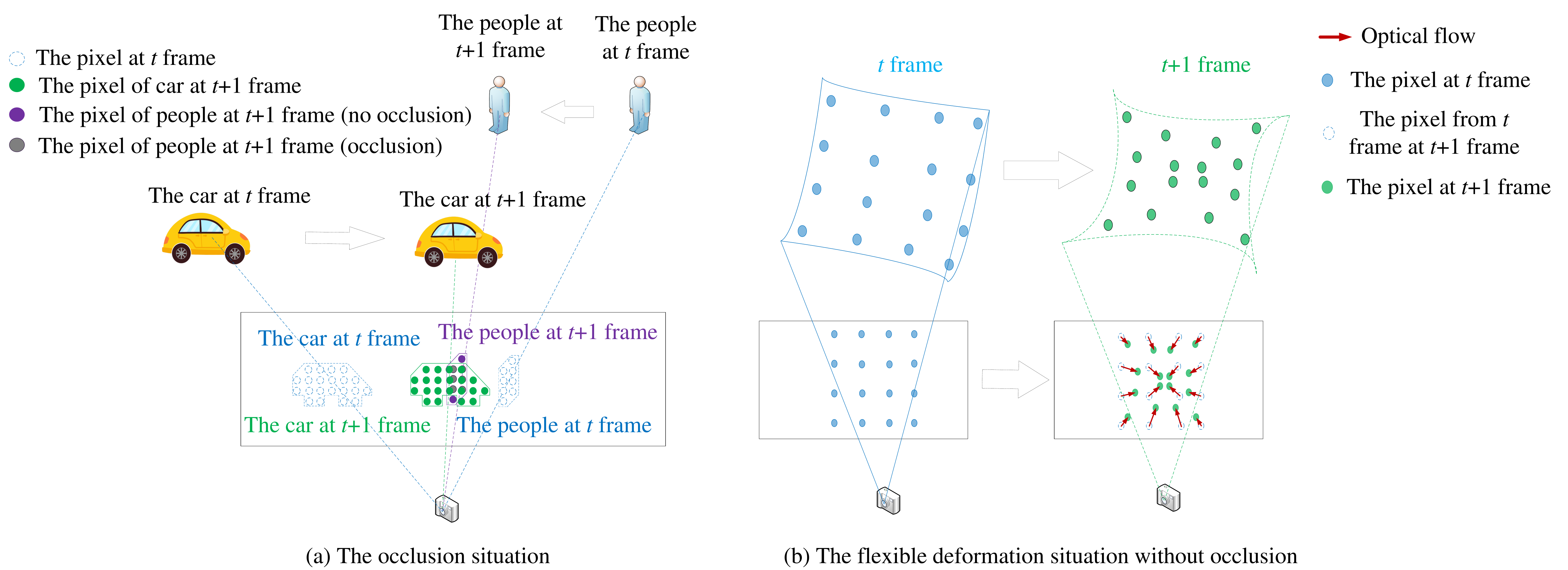}}
	\vspace{-3mm}
	\caption{Optical flow intersection and pixel blocking caused by occlusion. Fig. (a) shows the case of occlusion for rigid objects, while Fig. (b) shows the case of non-occlusion for flexible objects.}
	\label{occlusion_visual}
\end{figure*}

\begin{figure}[t]
	\centering
	\resizebox{1.00\columnwidth}{!}
	{
		\includegraphics[scale=1.00]{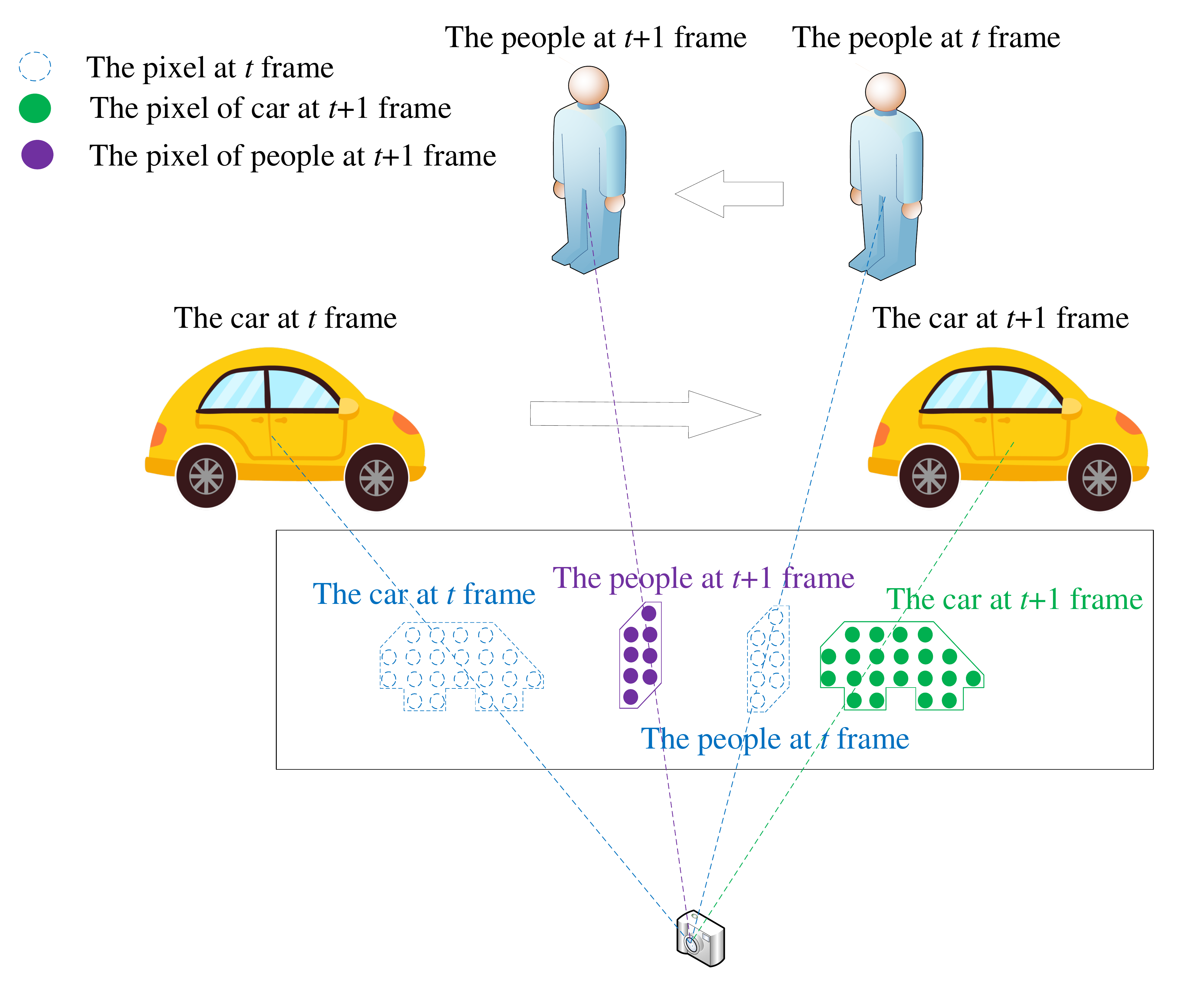}}
	\vspace{-4mm}
	\caption{An extreme situation that there are intersected flows, but there are not occlusion because of the big motion in consecutive frames.}
	\label{intersection}
	\vspace{-0pt}
\end{figure}

\begin{figure}[t]
	\centering
	\resizebox{0.80\columnwidth}{!}
	{
		\includegraphics[scale=1.00]{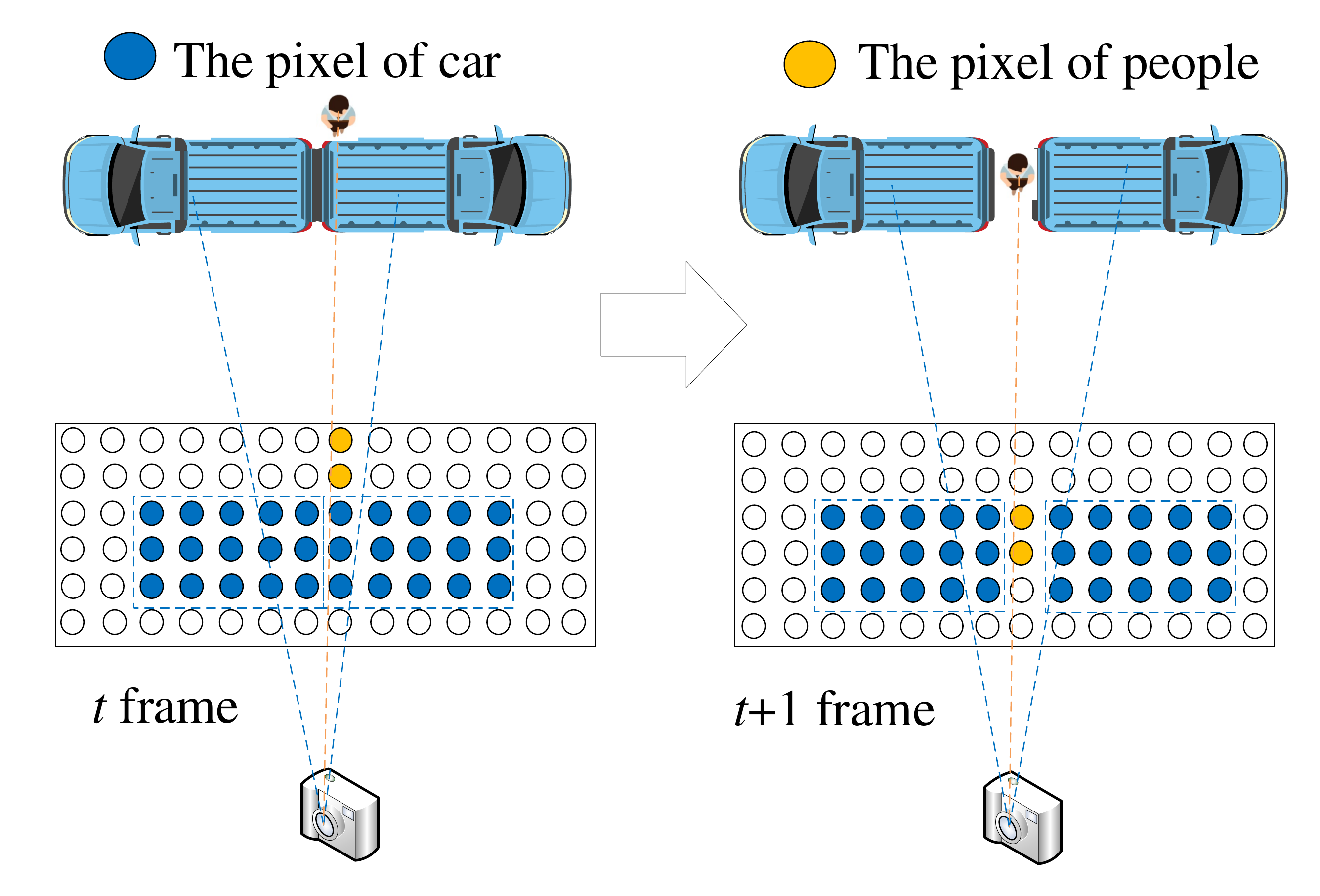}}
	\vspace{-0mm}
	\caption{An extreme situation that there are blocked pixels, but there are not occlusion because of the big motions of multiple objects in consecutive frames. }
	\label{encircling}
	\vspace{-0pt}
\end{figure}

\section{Related Work}

Optical flow describes the pixel displacement on a 2D projected image because of the relative 3D motion between objects and the camera for observing \cite{gibson1950perception}.
Traditional methods define optical flow estimation as an energy minimization problem based on brightness consistency and spatial smoothness \cite{horn1981schunck,brox2004high,sun2010secrets}. With the rapid development of deep learning, 
optical flow neural network can predict optical flow directly from a pair of images in an end-to-end manner \cite{dosovitskiy2015flownet,ilg2017flownet}. Ranjan et al. \cite{ranjan2017optical} propose the coarse-to-fine pyramid structure to make the network model size much smaller and improve the accuracy. Sun et al. \cite{ranjan2017optical} propose the PWC-Net, which performs warp operations and cost volume calculations for each level of the pyramid, showing the strong performance. Yang et al. \cite{yang2019volumetric} improve the volumetric layer by using the encoder-decoder architectures, to reduce parameters and achieve better performance. These supervised approaches need numerous data with optical flows labels to achieve better performance. However these data are expensive to obtain \cite{butler2012naturalistic,geiger2012we}, and sometimes special methods are even needed to get them, \cite{baker2011database}, which limits the application of these supervised methods.

The unsupervised approach avoids the need for labels through some regularization and has been the focus of recent research \cite{jason2016back,ren2017unsupervised,wang2018occlusion,meister2018unflow,alletto2018self,janai2018unsupervised,liu2019ddflow,liu2019selflow,zou2018df,ranjan2019competitive,wang2020unsupervised,jonschkowski2020matters}. The unsupervised method generates the optical flow by learning a function from the unlabeled dataset. As research goes on, the constraints of unsupervised training continue to increase, which allows neural networks to make more full use of unlabeled data, such as edge-aware smoothness \cite{wang2018occlusion}, photometric consistency loss \cite{wang2004image,sun2010secrets,zou2018df,meister2018unflow,ranjan2019competitive,liu2019ddflow,wang2019unos}, occlusion estimation \cite{wang2018occlusion,zou2018df,janai2018unsupervised,ranjan2019competitive,wang2020unsupervised}, distillation learning based on teacher and student models \cite{liu2019ddflow,liu2019selflow} and so on. 
UFlow \cite{jonschkowski2020matters} systematically compares those key components in an unsupervised optical flow model to identify which is most effective and choose the best combination of those components, achieving the better performance in all benchmarks.

Besides those key components of unsupervised optical flow estimation, there are many other improvements. Wang et al. \cite{wang2018occlusion} explicitly model occlusion and propose a new warping approach to solve the problem of large estimation errors caused by large motions. Alletto et al. \cite{alletto2018self} divide the optical flow estimation into two steps: global transformation with homography and refinement by a deeper network, which can make the optical flow estimation more accurate. Janai et al. \cite{janai2018unsupervised} firstly use multi-frame information for occlusion processing in the unsupervised learning of optical flow. SelFlow \cite{liu2019selflow} utilizes temporal information from multiple frames for better flow estimation. Zhong et al. \cite{zhong2019unsupervised} propose Deep Epipolar Flow which incorporates global geometric constraints into network learning. Flow2Stereo \cite{liu2020flow2stereo} trains a network to estimate both flow and stereo, using triangle constraint loss and quadrilateral constraint loss. Df-net \cite{zou2018df} proposes the cross consistency loss of the depth and pose based rigid flow and optical flow in rigid regions. Ranjan et al. \cite{ranjan2019competitive} bring forward the idea of competitive collaboration to achieve unsupervised coordinated training of four tasks: depth, camera motion, optical flow, and motion segmentation. Wang et al. \cite{wang2019unsupervised,wang2020unsupervised} jointly estimate pose, depth, and optical flow in an unsupervised method by dividing an image into three parts: the occluded region, the non-rigid region, and the rigid region. 

Many studied have done in these years for unsupervised learning of optical flow, as mentioned above. However, lots of works focus on the occlusion problem as the occlusion regions are not suitable for image reconstruction. There are seldom works on non-occlusion constraints. In this paper, novel unsupervised losses of optical flow are proposed based on geometric constraints in non-occlusion regions. The pixels in the non-occlusion regions are used to calculate these proposed losses: optical flow non-intersection loss and optical flow non-blocking loss, to punish the pixels that do not meet the constraints, which plays a guiding role in the model training.

\section{Geometric Laws of Optical Flow Field in the Non-occlusion Regions}\label{sec:laws}

2D image is a reflection of the real 3D world and the real motion takes place in 3D space. The 2D optical flow can be obtained by projecting the 3D scene flow to the 2D image plane as in Fig. \ref{occlusion_visual}. For the convenience of presentation and explanation, the camera is assumed to be stationary and the occlusion is caused by the motion of observed objects. In Fig. \ref{occlusion_visual}(a), at t frame, the car and the pedestrian can be seen by the camera, while the nearer car will occlude the farther pedestrian at $t+1$ frame. The pixels of cars and pedestrians at $t$ and $t+1$ frames are visualized on the image plane. The occluded pixels of the pedestrian will be surrounded by the pixels of the car. Similarly, the pixels of the car covering the pedestrian is also surrounded by the pixels of the pedestrian. At the same time, the pixels of different objects are intersected when occlusion appears. That is, flow intersection and pixel blocking have a connection with occlusion. Fig. \ref{occlusion_visual}(b) presents a non-occlusion flexible and deformable object. It can be seen that some pixels have a motion away from the camera in 3D space. There is an aggregated optical flow field but the optical flow is not intersected and the pixels are not blocked by surrounding adjacent pixel clusters.

From these observations, we infer the laws that the optical flow will not intersect each other and the pixels will not be blocked by surrounding adjacent pixel clusters in the non-occlusion regions. There are two extreme situations that are not consistent with the laws. It will be found that they happens so rarely in practice that the laws are satisfied in real applications.

As shown in Fig. \ref{intersection}, there is an extreme situation, where the car has a big motion and has intersected optical flows with the pedestrian, but they are not being occluded. They may accomplish the occlusion process in the consecutive frames or the trajectory of the car in the 2D image bypasses the pixels of the pedestrian. As in the consecutive frames, the motions are usually small compared with the size of the objects in the scene, our constraints are suitable in the real scenes.

If the pixels of something (eg. pedestrian in Fig. \ref{occlusion_visual}) are surrounded by the pixels of a nearer continuous body (eg. car in Fig. \ref{occlusion_visual}), the further thing will be occluded. However, if the surrounding four pixels are not from a continuous body, the assumption will not satisfied, as in Fig. \ref{encircling}. But the situation is so extreme that the inserted pedestrian and the two cars move so fast in the consecutive frames, and the pedestrian moves into the gap created by the movement of the cars.

\section{Unsupervised Learning of Optical Flow Based on Non-occlusion Constraints} \label{sec:constraints}

\subsection{The Overview of Our Unsupervised Framework of Optical Flow}  \label{sec:overview}

\begin{figure*}[t]
	\centering
	\resizebox{0.95\textwidth}{!}
	{
		\includegraphics[scale=1.00]{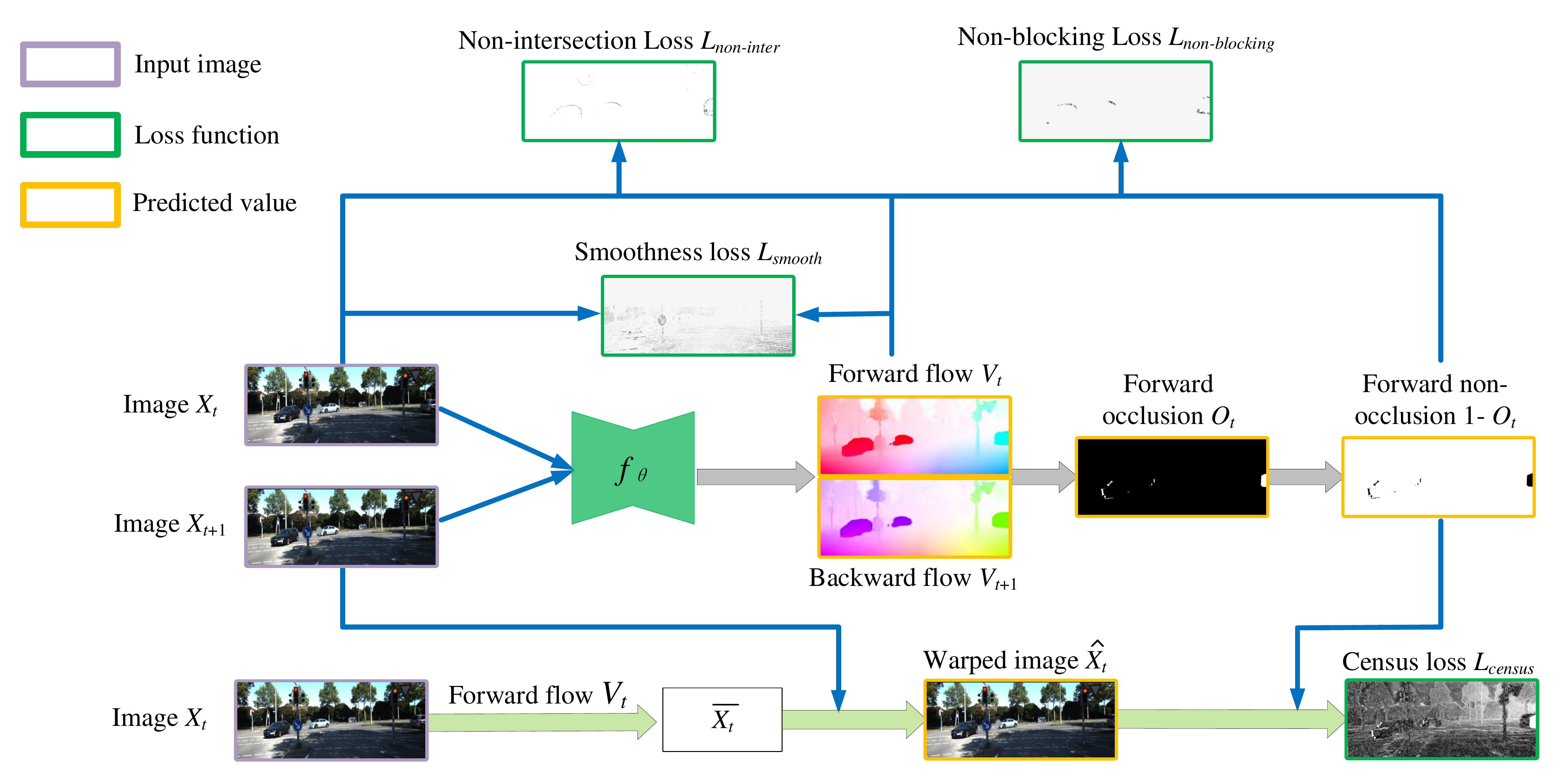}}
	\vspace{-0mm}
	\caption{The overview of our unsupervised learning pipeline of optical flow. Two images, $X_t$ and $X_{t+1}$, are fed into our network to estimate the optical flow. The forward-backward consistency based on the flow fields is used to estimate occlusion. The census loss compares the warped image $\widehat X_{t}$ to the corresponding original image $X_t$ and expresses their difference. Forward flow $V_t$ and backward flow $V_{t+1}$ are regularized using smoothness loss. Finally, the non-intersection loss and the non-blocking loss based on geometric constraints of optical flow are used to guide the training.}
	\label{network}
	\vspace{-0pt}
\end{figure*}

The overview of our unsupervised learning pipeline of optical flow is shown in Fig. \ref{network}. There are two adjacent images $X_t \in \mathbb{R}^{H \times W \times 3}$ and $X_{t+1} \in \mathbb{R}^{H \times W \times 3}$. They are input to an optical flow estimation network $f_\theta$ to get the forward optical flow $V_t = f_\theta(X_t,X_{t+1})$ and backward optical flow $V_{t+1} = f_\theta(X_{t+1},X_{t})$. The $V_t \in \mathbb{R}^{H \times W \times 2}$ indicates the 2D flow vector from $X_{t}$ to $X_{t+1}$ for each pixel in $X_t$, while $V_{t+1}$ indicates the optical flow from $X_{t+1}$ to $X_{t}$. Our objective is to obtain perfect parameters $\theta$ of the network from image sequences without the ground truth of optical flow to realize the optimized performance of optical flow. Fig. \ref{network} gives the losses in one direction ($t$ to $t+1$), and the other direction ($t+1$ to $t$) is similar. The consistency of forward and backward optical flow is used to estimate the occlusion regions \cite{wang2018occlusion}. Then, the non-occlusion regions are the other part in an image. 

The optical flow connects the images of adjacent frames at the pixel level. The optical flow can be unsupervised trained  by measuring the corresponding matching of the pixels between two frames. The idea of measuring pixel matching between adjacent frames is commonly realized through image warping \cite{ranjan2017optical,sun2018pwc}. Firstly, the corresponding coordinates after optical flow are calculated as: $[\hat{i},\hat{j}]^T=[i,j]^T + [u_t,v_t]^T$. Then, the warped image can be obtained by the differentiable bilinear interpolation: $\hat{X}_t(i,j)=\sum_{i\in{\lfloor \hat i \rfloor,\lceil \hat i \rceil},j\in{\lfloor \hat j \rfloor,\lceil \hat j \rceil}}w^{ij}X_t(i,j)$, $\sum_{i,j}w^{ij} = 1$. $\lceil \cdot \rceil$ means rounding up to ceil, and $\lfloor \cdot \rfloor$ means rounding down to floor. Then, census loss \cite{meister2018unflow} is used to enforce the consistency of warped image $\widehat X_{t}$ and original image $X_{t}$ as shown below:
\begin{equation}
\begin{split}
L_{census} =& \sum_{} ( 1 - O_t) \cdot  \sigma(\rho(X_t, \hat X_t))\\ + &\sum_{}(1-O_{t+1}) \cdot  \sigma(\rho(X_{t+1}, \hat X_{t+1})),
\end{split}
\end{equation}
where $O_t$ and $O_{t+1}$ are forward occlusion mask and backward occlusion mask, respectively. $\sigma\left(x\right) = \left( |x| + \epsilon \right)^q$ is the robust loss function \cite{liu2019ddflow}, where $\epsilon = 0.01$, $q = 0.4$. The brightness constancy $\rho (X_t,\hat X_t)$ is used to measure the difference between warped image and original image.

The smooth loss makes the estimated flow smooth according to the pixel gradient of the image. As with most methods, we use first-order and second-order smooth loss. The formula is shown as below:
 \begin{equation}
 L_{smooth\left(k\right)}=\frac{1}{N} \sum |\nabla^k V_t| \cdot  exp(-\frac{\mu}{3} \sum_{i}|\nabla X_t^i| ), 
 \end{equation}
where $\mu$ is weight based on the color channel ($i\in\{0,1,2\}$) of $X_t^i$ and $\mu = 150$. $k$ expresses the order of smoothness.

The non-intersection loss $L_{non-inter}$ and non-blocking $L_{non-block}$ loss are proposed in this paper to constraint and regulate optical flow learning inspired by the geometric laws of flow field introduced in Section \ref{sec:laws}. These losses are introduced in Section \ref{sec:intersection} and Section \ref{blocking}. In addition, we also utilize the idea of  distillation learning based on teacher and student models \cite{liu2019ddflow, liu2019selflow}. The loss of distillation learning is represented as $L_{dist}$.

In summary, the overall loss function is:
 \begin{equation}
 \begin{split}
L_{all} =& \alpha_1 L_{census} + \alpha_2 L_{smooth\left(k\right)} + \alpha_3 L_{non-inter} \\+& \alpha_4 L_{non-block} + \alpha_5 L_{dist},
\end{split}
 \end{equation}
where $\alpha_1 = 1$, $\alpha_2 = 4$, $\alpha_3 = \alpha_4 = 0.01$. As for the setting of $\alpha_5$, we follow the method of UFlow \cite{jonschkowski2020matters}, which is 0 for the first 50 percent of the training and then increases to a constant.

\subsection{The Non-intersection Loss} \label{sec:intersection}

Because the motion is small, the displacement described by optical flow can be regarded as the actual trajectories of pixel. Usually, an image has a big amount of pixels (eg. $448 \times 1024$ pixels in an image for the Sintel dataset \cite{butler2012naturalistic}), which is massive to calculate the relationship for the optical flow vectors of every two pixels. Inspired by the convolution in 2D, we calculated the results of each local regions and make all local calculation paralleled.

\begin{figure}[t]
	\centering
	\resizebox{0.90\columnwidth}{!}
	{
		\includegraphics[scale=1.00]{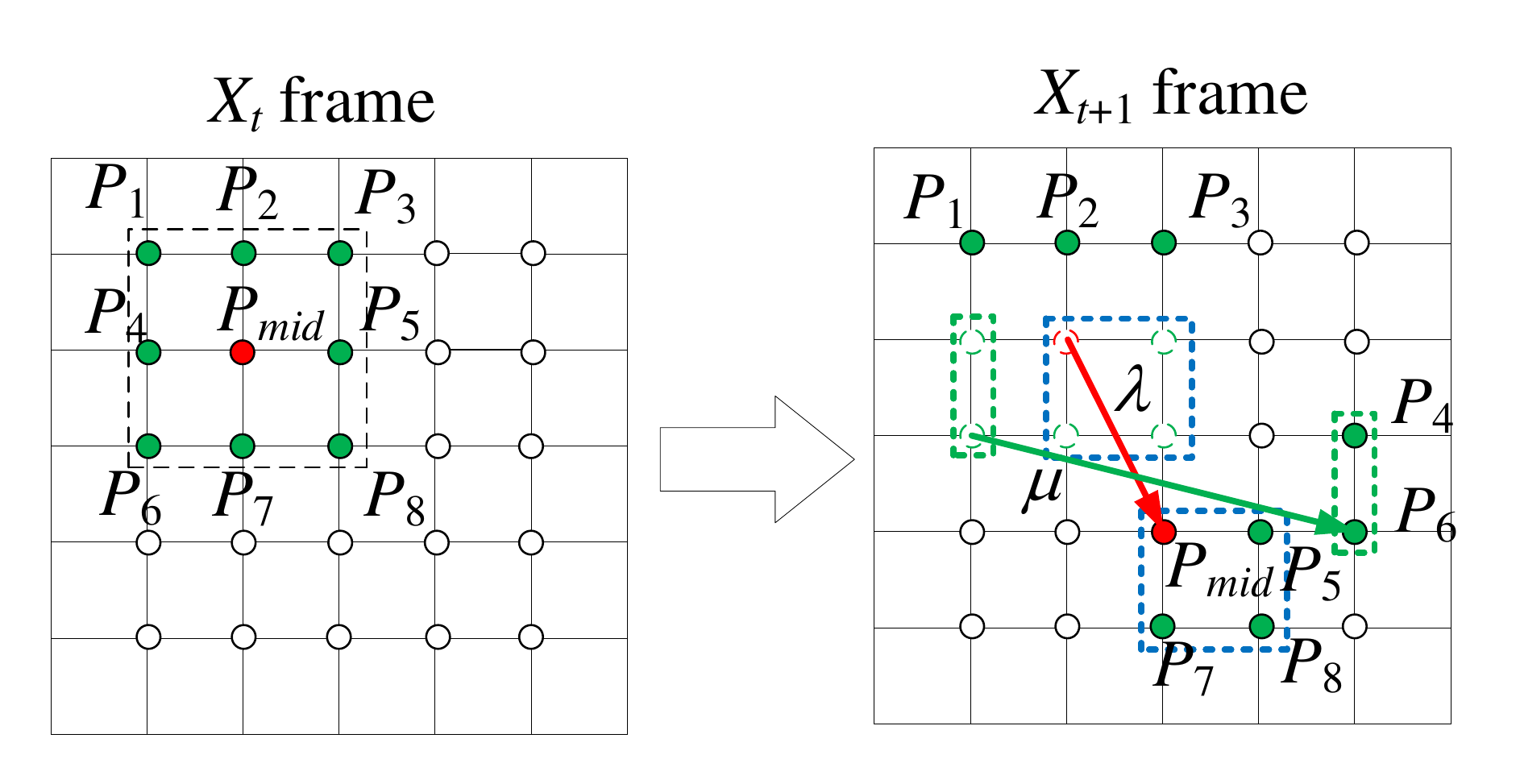}}
	\vspace{-1mm}
	\caption{Schematic diagram of the intersection of optical flow $\protect\overrightarrow{P_{mid}^t P_{mid}^{t+1}}$ and optical flow $\protect\overrightarrow{P_6^{t} P_6^{t+1}}$.}
	\label{non_intersection}
	\vspace{-0pt}
\end{figure}

A $3 \times 3$ sliding kernel is used to calculate the loss in each local area, and 1 is as the sliding step. For an image with a size of $H \times W$, there are $(H-2) \times (W-2)$ basic units totally. Fig. \ref{non_intersection} is a schematic diagram of the basic unit extracted by the sliding kernel, and the dashed box represents a basic unit extracted by the sliding kernel. With 1 as the sliding step, the sliding kernel moves one pixel to the right or to the down to extract the next basic unit. Based on this principle, $(H-2) \times (W-2)$ basic units with a size of $3 \times 3$ can be extracted from an image with a size of $H \times W$.

Parallel calculation is used to improve calculation efficiency. The whole image is divided into several basic units, and the optical flow non-intersection loss is calculated between the middle pixel and the 8 adjacent pixels in each basic unit. For the non-occlusion regions of the image $X_t$, as analysed in Section \ref{sec:laws}, the pixels will not intersect with each other, so the loss of optical flow non-intersection is calculated to penalize outliers. Among them, $P_{mid}$ corresponds to the middle pixel of a basic unit, and $P_i$ ($i=1,2,...8$) corresponds to other pixels adjacent to $P_{mid}$ in the basic unit. As shown in Fig. \ref{non_intersection}, the pixel points $P_i^t$ and $P_{mid}^{t}$ are in the first frame, and the corresponding pixels in the second frame are $P_i^{t+1}$ and $P_{mid}^{t+1}$. If the optical flow $\overrightarrow{P_{mid}^t P_{mid}^{t+1}}$ does not intersect the optical flow $\overrightarrow{P_i^{t} P_i^{t+1}}$, then the optical flow non-intersection loss between the pixel points $P_{mid}^t$ and $P_i^t$ is 0; Otherwise, it is considered that occlusion occurs between pixels. The optical flow non-intersection loss is calculated as the following steps.

There are usually different colors on both sides of the object edge, where the occlusion is prone to occur. The interior of the object is not easy to be occluded. Therefore, the loss function has a greater punishment on the interior of the object and less to the edge of the object. Therefore, the weight $w_i$ of optical flow non-intersection loss is calculated based on the degree of color difference between the middle pixel $P_{mid}$ and the surrounding 8 pixel points $P_i$ in each basic unit:
\begin{equation}{w_i} = \exp ( - \frac{1}{3}\sum\limits_{j \in \{ R,G,B\} } {\left| {{P_{i,j}} - {P_{mid,j}}} \right|} ),
\end{equation}
where $P_{mid}$ is the middle pixel of the basic unit, $P_i$ corresponds to other pixels adjacent to $P_{mid}$ in the basic unit ($i=1,2,...,8$). $P_{i,j}$ and $P_{mid,j}$ represent the three channel values of the RGB color space corresponding to the pixel points of $P_i$ and $P_{mid}$.

As shown in Fig. \ref{non_intersection}, the optical flow intersection coefficients $\mu_i$ and $\lambda_i$ represent the ratio of the intersection position to the length of the optical flow vector:
\begin{equation}\begin{gathered}
		{\lambda _i} = \frac{1}{\Lambda }\left| {\begin{array}{*{20}{c}}
				{{x_i^t} - {x_{mid}^t}}&{ - \Delta {x_i^t}} \\ 
				{{y_i^t} - {y_{mid}^t}}&{ - \Delta {y_i^t}} 
		\end{array}} \right|, \hfill \\
		{\mu _i} = \frac{1}{\Lambda }\left| {\begin{array}{*{20}{c}}
				{\Delta {x_{mid}^t}}&{{x_i^t} - {x_{mid}^t}} \\ 
				{\Delta {y_{mid}^t}}&{{y_i^t} - {y_{mid}^t}} 
		\end{array}} \right|, \hfill \\ 
	\Lambda  =  - \Delta {x_{mid}^t}\Delta {y_i^t} + \Delta {x_i^t}\Delta {y_{mid}^t},
	\end{gathered}
\end{equation}
where $(x_{mid}^t, y_{mid}^t)$ is the coordinate of the intermediate pixel $P_{mid}^t$ in the $t$ frame, and $(x_i^t, y_i^t)$ is the coordinate of adjacent pixel $P_{i}^t$ in the $t$ frame.  $\overrightarrow{P_{mid}^t P_{mid}^{t+1}} = (\Delta {x_{mid}^t}, \Delta {y_{mid}^t})$ and $\overrightarrow{P_{i}^t P_{id}^{t+1}} = (\Delta {x_{i}^t}, \Delta {y_{i}^t})$ are the optical flow displacements of the pixel points, $P_{mid}^t$ and $P_{i}^t$ , from the $t$ frame to the $t+1$ frame. When $0<\mu_i,\lambda_i<1$, $\overrightarrow{P_{mid}^t P_{mid}^{t+1}}$ and $\overrightarrow{P_{i}^t P_{id}^{t+1}}$ will intersect.

The optical flow non-intersection loss $L_k$ of the intermediate pixel relative to all 8 adjacent surrounding pixels in the $k$-th basic unit is calculated as follows:
\begin{equation}
{L_k} = \left\{ {\begin{array}{*{20}{l}}
		{\frac{1}{8}\sum\limits_{i = 1}^8 {{w_i}\sigma\left(\exp ( - {{\left| {{\lambda _i} - {\mu _i}} \right|}^2})\right)},}&{0 < {\lambda _i},{\mu _i} < 1,} \\ 
		{0,}&{{\text{Others}}.} 
\end{array}} \right.
\end{equation}

$L_k$ is calculated in one basic unit and a total of $(H-2)\times(W-2)$ basic units are extracted. Then, $L_{non-inter}$ represents the average non-intersection loss of optical flow for these $(H-2)\times(W-2)$ units:
 \begin{equation}
  L_{non-inter} = \frac{1}{(H-2)\times(W-2)}\sum\limits_{k = 1}^{(H-2)\times (W-2)} {L_k}. \end{equation}

\subsection{The Non-blocking Loss}\label{blocking}

As analysed in Section \ref{sec:laws}, in the non-occlusion area, there will be no pixel blocking when the object moving. The parallelization technique is also used here similar to Section \ref{sec:intersection}.

\begin{figure}[t]
	\centering
	\resizebox{0.95\columnwidth}{!}
	{
		\includegraphics[scale=1.00]{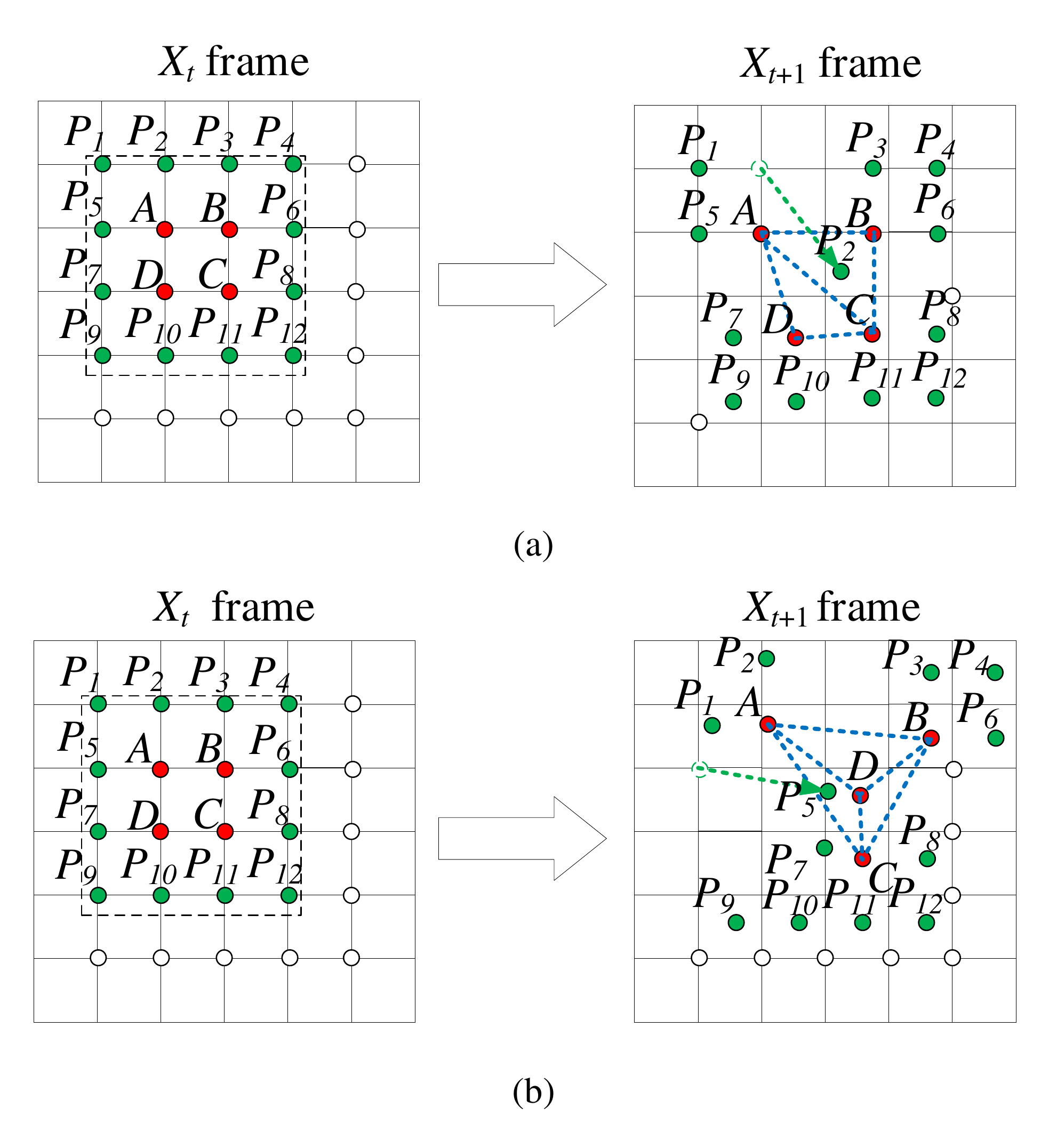}}
	\vspace{-3mm}
	\caption{The geometric relation diagram of peripheral  pixels and quadrilateral $\overline {ABCD}$. (a) corresponds to the situation that quadrilateral $\overline {ABCD}$ is convex quadrilateral, and (b) corresponds to the situation that quadrilateral $\overline {ABCD}$ is concave quadrilateral. }
	\label{non_blocking}
	\vspace{-0pt}
\end{figure}

Taking $4\times 4$ as the sliding kernel size and 1 as the sliding step size. Fig. \ref{non_blocking} shows the schematic diagram of the pixel blocking for a basic unit extracted by the sliding kernel, and the dashed box represents a basic unit extracted by the sliding kernel. With 1 as the sliding step, the sliding kernel moves one pixel to the right or to the down to extract the next basic unit. Based on this principle, $(H-3) \times (W-3)$ basic units with a size of $4 \times 4$ can be extracted from the image at $t$ frame with a size of $H \times W$. Parallelization is used to calculate the extracted $(H-3) \times (W-3)$ basic units. In each basic unit, the optical flow non-blocking loss of 12 pixels in the periphery is calculated based on the blocking calculation with the 4 pixels in the middle. Our proposed method determines and measures how far the surrounding pixels flow into the inside of the quadrilateral composed of 4 pixels in the middle. 

Define a basic unit with 4 pixels ${A,B,C,D}$ in the middle and 12 pixels $P_i$ in the periphery, where $i=1,2,...,12$. The four pixels $A, B, C, D$ in the middle of the basic unit at $t$ frame constitute a quadrilateral $\overline {ABCD}$ at $t+1$ frame, which can be divided into two triangles by a diagonal line. When the quadrilateral $\overline {ABCD}$ at the second frame is a convex quadrilateral, according to the selection of different diagonals $\overline {AC}$ or $\overline {BD}$ for division, there are two cases where the quadrilateral $\overline {ABCD}$ contains two triangles. For any of the two division cases, if the peripheral pixels flow into a triangle, it can be inferred that occlusion occurs. (For the sake of brief expression, we classify the points falling on the boundary of the triangle as being within the triangle, because this will not affect the subsequent distance calculation.) As shown in Fig. \ref{non_blocking}(a), According to the diagonal $\overline {AC}$, the quadrilateral is divided into triangles $\overline {ABC}$ and $\overline {ACD}$. The point $P_i$ flows inside triangle $\overline {ABC}$.

However, if the quadrilateral $\overline {ABCD}$ at the second frame is a concave quadrilateral, the division case with diagonals $\overline {AC}$ as shown in Fig. \ref{non_blocking}(b) is not enough to determine that the pixel $p_5$ falls inside the quadrilateral $\overline {ABCD}$. If $\overline {AC}$ is as the diagonal in the calculation progress, the point $P_5$ is simultaneously inside triangles $\overline {ABC}$ and $\overline {ACD}$. However, the point $P_5$ is not blocked by the quadrilateral $\overline {ABCD}$, but this can be judged by using the diagonal $\overline {BD}$ to divide. According to the diagonal $\overline {BD}$, the quadrilateral $\overline {ABCD}$ is divided to triangles $\overline {ABD}$ and $\overline {BCD}$, and the point $P_5$ is not inside $\overline {ABD}$ or $\overline {BCD}$. Therefore, only the point $P_i$ is inside a triangle both in the two division cases, the pixel blocking occurs for the point $P_i$.

For triangle $\overline {ABC}$ and point $P_i$, when $\overrightarrow{BA} \times \overrightarrow{BP}, \overrightarrow{AC} \times \overrightarrow{AP}, \overrightarrow{CB} \times \overrightarrow{CP}$ are in the same direction, it can be judged that $P_i$ is within $\overline {ABC}$. In the same way, it can be inferred if $P_i$ is within ${\overline{ACD}}$, $\overline{ABD}$ and $\overline{BCD}$. The logic expression is as follows:
\begin{small}\begin{equation}\begin{gathered}
			{\Gamma _{\overline {ABC} }} =\!\! \left( {\left|\!\!\! {\begin{array}{*{20}{c}}
						{{x_{\overrightarrow {BA} }}}\!\!\! &\!\! {{x_{\overrightarrow {BP} }}} \\ 
						{{y_{\overrightarrow {BA} }}}\!\!\! &\!\! {{y_{\overrightarrow {BP} }}} 
				\end{array}} \!\!\!\right| \!\! \geqslant \!\! 0} \right)\!\! \wedge \!\! \left( {\left|\!\!\! {\begin{array}{*{20}{c}}
						{{x_{\overrightarrow {AC} }}}\!\!\! &\!\! {{x_{\overrightarrow {AP} }}} \\ 
						{{y_{\overrightarrow {AC} }}}\!\!\! &\!\! {{y_{\overrightarrow {AP} }}} 
				\end{array}} \!\!\!\right| \!\!\geqslant\!\! 0} \right) \!\!\wedge\!\! \left( {\left|\!\!\! {\begin{array}{*{20}{c}}
						{{x_{\overrightarrow {CB} }}}\!\!\! &\!\! {{x_{\overrightarrow {CP} }}} \\ 
						{{y_{\overrightarrow {CB} }}}\!\!\! &\!\! {{y_{\overrightarrow {CP} }}} 
				\end{array}} \!\!\! \right| \!\!\geqslant\!\! 0} \right) \hfill \\
			{\Gamma _{\overline {ACD} }} =\!\! \left( {\left|\!\!\! {\begin{array}{*{20}{c}}
						{{x_{\overrightarrow {CA} }}}\!\!\! &\!\! {{x_{\overrightarrow {CP} }}} \\ 
						{{y_{\overrightarrow {CA} }}}\!\!\! &\!\! {{y_{\overrightarrow {CP} }}} 
				\end{array}} \!\!\!\right| \!\!\geqslant\!\! 0} \right) \!\!\wedge\!\! \left( {\left|\!\!\! {\begin{array}{*{20}{c}}
						{{x_{\overrightarrow {AD} }}}\!\!\! &\!\! {{x_{\overrightarrow {AP} }}} \\ 
						{{y_{\overrightarrow {AD} }}}\!\!\! &\!\! {{y_{\overrightarrow {AP} }}} 
				\end{array}} \!\!\!\right| \!\!\geqslant\!\! 0} \right) \!\!\wedge\!\! \left( {\left|\!\!\! {\begin{array}{*{20}{c}}
						{{x_{\overrightarrow {DC} }}}\!\!\! &\!\! {{x_{\overrightarrow {DP} }}} \\ 
						{{y_{\overrightarrow {DC} }}}\!\!\! &\!\! {{y_{\overrightarrow {DP} }}} 
				\end{array}} \!\!\!\right| \!\!\geqslant\!\! 0} \right) \hfill \\
			{\Gamma _{\overline {ABD} }} =\!\! \left( {\left|\!\!\! {\begin{array}{*{20}{c}}
						{{x_{\overrightarrow {BA} }}}\!\!\! &\!\! {{x_{\overrightarrow {BP} }}} \\ 
						{{y_{\overrightarrow {BA} }}}\!\!\! &\!\! {{y_{\overrightarrow {BP} }}} 
				\end{array}} \!\!\!\right| \!\!\geqslant\!\! 0} \right) \!\!\wedge\!\! \left( {\left|\!\!\! {\begin{array}{*{20}{c}}
						{{x_{\overrightarrow {AD} }}}\!\!\! &\!\! {{x_{\overrightarrow {AP} }}} \\ 
						{{y_{\overrightarrow {AD} }}}\!\!\! &\!\! {{y_{\overrightarrow {AP} }}} 
				\end{array}} \!\!\!\right| \!\!\geqslant\!\! 0} \right) \!\!\wedge\!\! \left( {\left|\!\!\! {\begin{array}{*{20}{c}}
						{{x_{\overrightarrow {DB} }}}\!\!\! &\!\! {{x_{\overrightarrow {DP} }}} \\ 
						{{y_{\overrightarrow {DB} }}}\!\!\! &\!\! {{y_{\overrightarrow {DP} }}} 
				\end{array}} \!\!\!\right| \!\!\geqslant\!\! 0} \right) \hfill \\
			{\Gamma _{\overline {BCD} }} =\!\! \left( {\left|\!\!\! {\begin{array}{*{20}{c}}
						{{x_{\overrightarrow {BD} }}}\!\!\! &\!\! {{x_{\overrightarrow {BP} }}} \\ 
						{{y_{\overrightarrow {BD} }}}\!\!\! &\!\! {{y_{\overrightarrow {BP} }}} 
				\end{array}} \!\!\!\right| \!\!\geqslant\!\! 0} \right) \!\!\wedge\!\! \left( {\left|\!\!\! {\begin{array}{*{20}{c}}
						{{x_{\overrightarrow {DC} }}}\!\!\! &\!\! {{x_{\overrightarrow {DP} }}} \\ 
						{{y_{\overrightarrow {DC} }}}\!\!\! &\!\! {{y_{\overrightarrow {DP} }}} 
				\end{array}} \!\!\!\right| \!\!\geqslant\!\! 0} \right) \!\!\wedge\!\! \left( {\left|\!\!\! {\begin{array}{*{20}{c}}
						{{x_{\overrightarrow {CB} }}}\!\!\! &\!\! {{x_{\overrightarrow {CP} }}} \\ 
						{{y_{\overrightarrow {CB} }}}\!\!\! &\!\! {{y_{\overrightarrow {CP} }}} 
				\end{array}} \!\!\!\right| \!\!\geqslant\!\! 0} \right) \hfill 
\end{gathered} \end{equation}\end{small}
where $\Gamma_{\overline {ABC} },\Gamma_{\overline {ACD} },\Gamma_{\overline {ABD} }$, and $\Gamma_{\overline {BCD} }$ represent if $P_i$ is in the triangles $\overline {ABC}, \overline {ACD},\overline {ABD}$, and $\overline {BCD}$, respectively. As analysed above, it is inferred that $P_i$ flows into  the quadrilateral $\overline {ABCD}$ when $P_i$ flows at least into a triangle both in the two division cases, that is:
\begin{equation}
	{\Gamma _{\overline {ABCD} }}=\left( {{\Gamma _{\overline {ABC} }} \vee {\Gamma _{\overline {ACD} }}} \right) \wedge \left( {{\Gamma _{\overline {ABD} }} \vee {\Gamma _{\overline {BCD} }}} \right).
\end{equation}

According to the spatial geometric relationship between the quadrilateral $\overline {ABCD}$ formed by the intermediate four pixels and each peripheral pixel $P_i$ in a basic unit, the optical flow non-blocking loss $E_i$ of the pixel $P_i$ is defined as:
\begin{equation}
{{E_k}} = \left\{ {\begin{array}{*{20}{l}}
			{\frac{1}{{12}}\sum\limits_{i = 1}^{12}{e^{ - \frac{1}{{{d_i}}}}},}&{{\Gamma _{\overline {ABCD} }} = True,} \\ 
			{0,}&{{\Gamma _{\overline {ABCD} }} = False,}
	\end{array}} \right.\end{equation}
where $d_i$ is the minimum distance of $P_i$ to each side of the quadrilateral $\overline {ABCD}$.

The above is about the loss for a peripheral pixel $P_i$ in a basic unit. There are a total of $(H-3)\times (W-3)$ basic units for the source image, and each unit includes $12$ peripheral pixels. The optical flow  non-blocking loss of the entire image is as follows:
\begin{equation}
{L_{{\text{non  -  blocking }}}} = \frac{1}{{(H - 3)\times(W - 3)}}\sum\limits_{k = 1}^{(H - 3)\times(W - 3)} {{E_k}}.
\end{equation}

\setlength{\tabcolsep}{6mm}
\begin{table*}[t]
	\centering
		\caption{The experiment results training and testing on synthetic dataset. 'ft' means that the models of optical flow estimation are fine-tuned on specific evaluation datasets after supervising, which is not conducive to practical application. The results in parentheses are not comparable because its training set contains the evaluation set. "()" means that the results are supervised trained on the evaluation set. "\{\}" means that the results are unsupervised trained on the evaluation set, and "[]" means that the results are trained on the dataset related to the evaluation set. The best results under every evaluation set are marked in bold. Unpublished results are marked as '-'.}
	\footnotesize
	\begin{center}
		
		\begin{tabular}{lccccc}
			\toprule
			&&\multicolumn{2}{c}{EPE on Sintel Clean\cite{butler2012naturalistic}}  &\multicolumn{2}{c}{EPE on Sintel Final\cite{butler2012naturalistic}}       \\ 
			\cline{3-6}\noalign{\smallskip}
			\multirow{-2}{*}{\begin{tabular}[c]{@{}c@{}} Method\end{tabular}}

			&\multirow{-2}{*}{\begin{tabular}[c]{@{}c@{}} Multi-frame\end{tabular}}
			& Train & Test & Train & Test \\
			
			\hline
			\noalign{\smallskip}
			
			FlowNet2-ft \cite{ilg2017flownet}
			&
			&(1.45)	&4.16
			&(2.01)	&5.74

			\\                		
			PWC-Net-ft \cite{sun2018pwc} 
			&
			&(1.70)	&3.86
			&(2.21)	&5.13

			\\
			SelFlow-ft \cite{liu2019selflow}   
			&
			&(1.68)	&[3.74]
			&(1.77)	&\{4.26\}

			\\
			VCN-ft \cite{yang2019volumetric}  
			&
			&(1.66)	&2.81
			&(2.24)	&4.40

			\\
			\hline
			\noalign{\smallskip}
			FlowNet2 \cite{ilg2017flownet}  
		    &
			&\bf2.02	&\bf3.96
			&\bf3.14	&\bf6.02

			\\
			PWC-Net \cite{sun2018pwc}  
			&
			&2.55	&-
			&3.93	&

			\\
			VCN \cite{yang2019volumetric}  
			&
			&2.21	&-
			&3.62	&-
            
			\\
			\hline
			\noalign{\smallskip}
			DSTFlow \cite{ren2017unsupervised}  
			&
			&\{6.16\}	&10.41
			&\{7.38\}	&11.28

			\\
			OAFlow \cite{wang2018occlusion}   
			&
			&\{4.03\}	&7.95
			&\{5.95\}	&9.15

			\\
			UnFlow \cite{meister2018unflow}  
			&
			&-      &-
			&7.91	&10.21

			\\
			MFOccFlow \cite{janai2018unsupervised}
			&$\surd$
			&\{3.89\}	&7.23
			&\{5.52\}	&8.81
			\\
			EPIFlow \cite{zhong2019unsupervised}
			&$\surd$
			&3.94	&7.00
			&5.08	&8.51
			\\
			DDFlow \cite{liu2019ddflow}   
			&$\surd$
			&\{2.92\}&6.18
			&\{3.98\}	&7.40
			\\
			SelFlow \cite{liu2019selflow}
			&$\surd$
			&[2.88]	&[6.56]
			&\{3.87\}	&\{6.57\}
			\\
			UFlow-test\cite{jonschkowski2020matters}  
			&
			&3.01	&-
			&4.09	&-
			\\
			UFlow-train\cite{jonschkowski2020matters}  
			&
			&\{2.50\}	&5.21
			&\{3.39\}	&6.50
			\\
			Our-test   
			&
			&\bf2.94	&-
			&\bf3.95	&-
			\\
			Our-train  
			&
			&\{2.47\}	&\bf4.26
			&\{3.57\}	&\bf6.28
			\\
			\bottomrule
		\end{tabular}
		
	\end{center}
	\vspace{-0pt}
	\label{sintel}
\end{table*}

\setlength{\tabcolsep}{4.5mm}
\begin{table*}[t]
	\centering
		\caption{The experiment results training on the synthetic datasets while testing on both synthetic and real datasets. 'MF' means that the method utilizes the information of multiple frames. The best results for each evaluation dataset are marked in bold. }
	\footnotesize
	\begin{center}
		
		\begin{tabular}{clcccccc}
			\toprule
			&&\multicolumn{1}{c}{Flying Chairs \cite{dosovitskiy2015flownet}}  &\multicolumn{2}{c}{Sintel train \cite{butler2012naturalistic}}  	&\multicolumn{3}{c}{KITTI-2015 train \cite{menze2015joint}}
			\\ 
			\cline{3-8}\noalign{\smallskip}

			\multirow{-2}{*}{\begin{tabular}[c]{@{}c@{}} Training Dataset\end{tabular}}
			&\multirow{-2}{*}{\begin{tabular}[c]{@{}c@{}} Method \end{tabular}}& Test & Clean & Final & All &Noc &ER\% \\
			
			\hline
			\noalign{\smallskip}
			\multirow{5.5}{*}{\begin{tabular}[c]{@{}c@{}} Flying Chairs \cite{dosovitskiy2015flownet} \\dataset \end{tabular}}
			&PWC-Net \cite{sun2018pwc}      
			&\bf2.00	&3.33
			&4.59	&13.20
			&-      &41.79

			\\    
			&DDFlow \cite{liu2019ddflow} {(MF)}     
			&2.97   &4.83
			&4.85	&17.26
			&-      &-
			\\
			&UFlow-test \cite{jonschkowski2020matters}     
			&\{2.82\}	&4.36
			&5.12	&15.68
			&7.96      &32.69
			\\
			&UFlow-train \cite{jonschkowski2020matters}     
			&2.55	&3.43
			&4.17	&11.27
			&5.66   &30.31
			\\
			&Ours-test    
			&2.72	&4.63
			&5.32	&15.43
			&7.23   &28.11
			\\
			&Ours-train    
			&2.52	&\bf3.23
			&\bf4.15	&\bf9.84
			&\bf4.65   &\bf26.67
			\\

			\hline
			\noalign{\smallskip}
			\multirow{5.5}{*}{\begin{tabular}[c]{@{}c@{}}Sintel  \\dataset \cite{butler2012naturalistic}\end{tabular}}
			&PWC-Net\cite{sun2018pwc}      
			&3.69	&\bf(1.86)
			&\bf(2.31)	&10.52
			&-      &30.49

			\\    
			&DDFlow \cite{liu2019ddflow} (MF)     
			&3.46   &\{2.92\}
			&\{3.98\}	&12.69
			&-      &-
			\\
			&UFlow-test \cite{jonschkowski2020matters}     
			&3.39	&3.01
			&4.09	&7.67
			&3.77      &\bf17.41
			\\
			&UFlow-train \cite{jonschkowski2020matters}     
			&3.25	&\{2.50\}
			&\{3.39\}	&9.40
			&4.53   &20.02
			\\
			&Ours-test    
			&2.87	&2.94
			&3.95	&\bf7.52
			&\bf3.38   &18.63
			\\
			&Ours-train    
			&\bf2.82	&\{2.47\}
			&\{3.57\}	&8.40
			&3.60   &20.27
			\\

			\bottomrule
		\end{tabular}
		
	\end{center}
	\vspace{-0pt}
	\label{KITTI}
\end{table*}

\section{Experiments}
\subsection{Training and Testing Details}\label{detail}
In order to demonstrate the effectiveness of our proposed method, our model is evaluated on the standard optical flow benchmark datasets: Flying Chairs dataset \cite{dosovitskiy2015flownet}, Sintel dataset \cite{butler2012naturalistic}, and KITTI 2015 datasets \cite{geiger2012we,menze2015joint}. Flying Chairs and Sintel are synthetic datasets, and KITTI is a real dataset. The Flying Chairs dataset contains a total of 22,872 pairs of images, of which 22,232 pairs are used as the training set and the remaining 640 pairs are used as the test set. For the Sintel dataset, we divide the training set and test set according to the standard classification criteria, where the training set contains 2082 images and the test set contains 1128 images. The training set and test set in KITTI 2015 dataset both contain 200 pairs of images. For Sintel, it is  common to train on the training set, and report the benchmark performance on the test set, which is included in our experiment. We expect to evaluate the generalization ability of our model on different datasets. However, the test set does not have public labels and there is a limit on the number of submissions to the official test set, so to be convenient for our experiments, we also train on the test set and evaluate on the training set. Therefor, there are two trained models for Sintel dataset. One is trained on the training set, the other is trained on the test set. Since the Sintel dataset contains both final and clean parts, they are used both when training the model and separately when evaluating the model, like UFlow \cite{jonschkowski2020matters}. In addition, pretraining is a very common method to improve accuracy in both supervised\cite{dosovitskiy2015flownet,sun2018pwc} and unsupervised\cite{liu2019ddflow,zhong2019unsupervised} optical flow estimation, so we have a pretraining stage in the trainging set of Flying Chairs before our formal training on Sintel. For KITTI, the raw KITTI 2015 dataset is used to verify the generalization ability of our model. We use the training set of KITTI 2015 as our evaluation set because the training set has public ground truth of the optical flow. It is expected that our method can be trained on synthetic datasets and evaluated on real datasets to achieve better generalization performance.

\begin{figure*}[t]
	\centering
	\resizebox{1.00\textwidth}{!}
	{
		\includegraphics[scale=1.00]{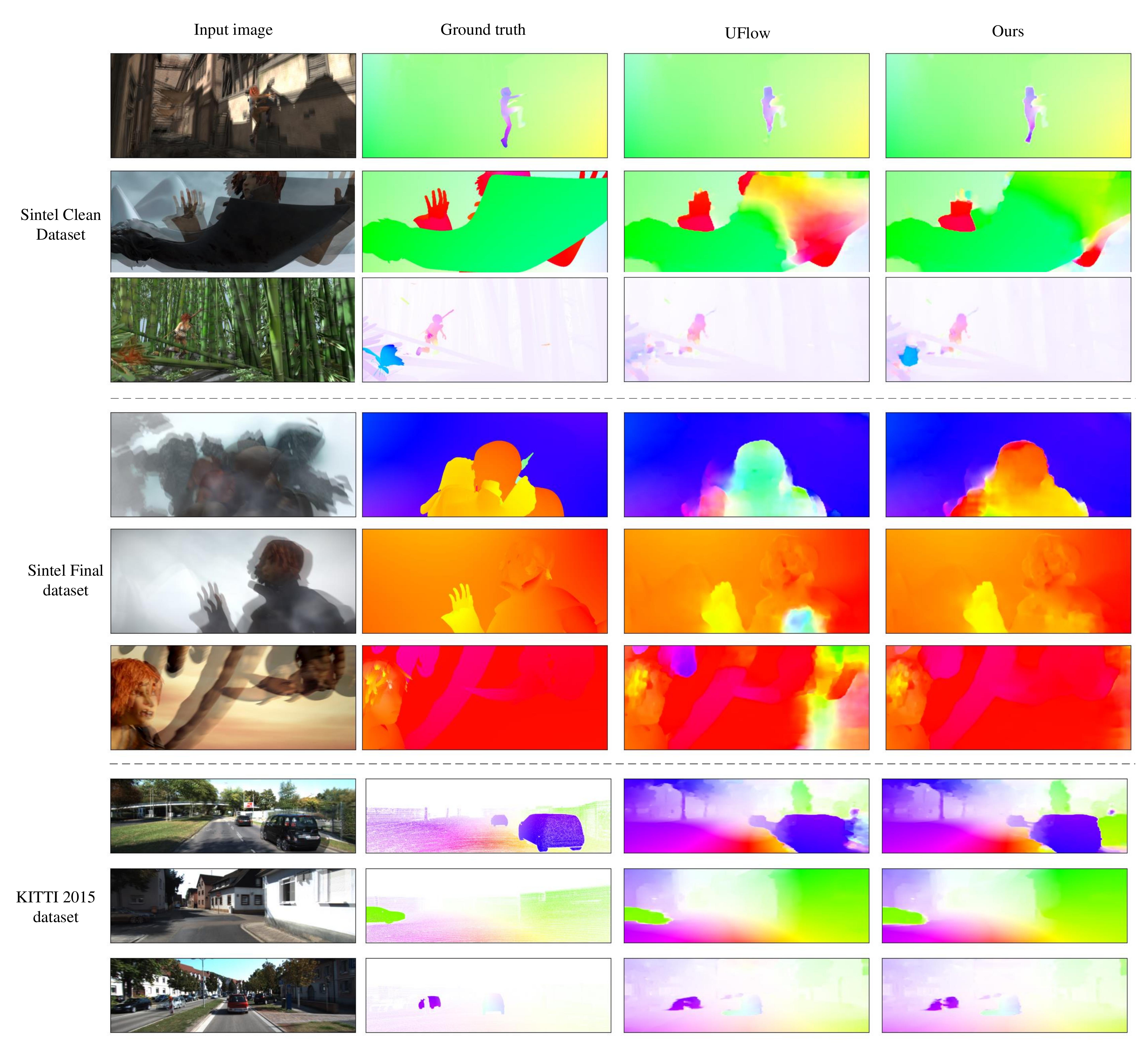}}
	\vspace{-4mm}
	\caption{The visualization of our qualitative results compared with UFlow \cite{jonschkowski2020matters} on Sintel Clean, Sintel Final, and KITTI 2015 datasets respectively. These qualitative results show the effect of our method, with good results for independent moving objects as well as fine details.}
	\label{vision}
	\vspace{-0pt}
\end{figure*}

As with most previous works, we use EPE (Endpoint Error) and ER (Error Rates) as our evaluation metrics. Our network structure is based on PWC-Net \cite{sun2018pwc}, similar to UFlow \cite{jonschkowski2020matters}. All experiments are performed with 1 as the batch size on a single of RTX 2080Ti. Our experiments are based on TensorFlow 2.2.0. The Adam \cite{kingma2014adam} method is used as the optimizing strategy of training, where $\beta_1 = 0.9$, $\beta_2 = 0.999$. In the pretraining stage, the learning rate is a constant, $10^{-4}$. In the training phase, exponential decay is used. The learning rate decays 0.5 times per 200 epochs from $10^{-4}$ to $10^{-8}$, similar to UFlow \cite{jonschkowski2020matters}.

\subsection{Results}

Two series of experiments are conducted. The first group of experiments are trained and evaluated on the synthetic dataset. The second group of experiments are trained on synthetic dataset and evaluated on real dataset to verify the generalization ability of our method. Experimental methods and details are as Section \ref{detail}.

\subsubsection{Testing on Synthetic Dataset}

The quantitative evaluation results on the synthetic dataset (Sintel dataset) are shown in Table \ref{sintel}, which shows the results of unsupervised and supervised optical flow methods. Compared with MFOccFlow \cite{janai2018unsupervised}, EPIFlow \cite{zhong2019unsupervised}, DDFlow \cite{liu2019ddflow} and SelFlow \cite{liu2019selflow}, our method does not need to use the information of multiple frames but achieves better performance, which depends on our proposed novel constraints in the non-occlusion regions. In addition, SelFlow \cite{liu2019selflow} downloads the raw Sintel movie and extracts about 10,000 images, which makes its training data include both the official training set and the test set, while our method only trains on the official training set or test set. UFlow \cite{jonschkowski2020matters} systematically compares and improves occlusion segmentation methods. On the basis of getting the fine non-occlusion regions, we propose a geometry-based unsupervised constraint method for the optical flow in the non-occlusion regions. Compared with UFlow \cite{jonschkowski2020matters}, we make better use of the details of the non-occlusion regions and achieve better performance.

\subsubsection{Generalization Test on Real Dataset}

We evaluate the generalization of the model on the real dataset, KITTI. To evaluate the generalization of the model broadly, the models trained on different datasets will be evaluated on all datasets, and the results are shown in Table \ref{KITTI}. Specifically, models are trained on synthetic datasets, Flying Chairs and Sintel, and tested on multiple datasets, Flying Chairs, Sintel, and KITTI. Compared with PWC-Net \cite{sun2018pwc}, our unsupervised results are not as good as that of PWC-Net \cite{sun2018pwc} when trained and tested on the Flying Chairs dataset. However, our generalization performance outperforms the supervised approach, PWC-Net \cite{sun2018pwc}, on both the more complex synthetic data, Sintel, as well as the real dataset, KITTI. When trained on Sintel dataset, our generalization performance on both Flying Chairs and KITTI also outperforms PWC-Net \cite{sun2018pwc}. We use the network structure similar to PWC-Net \cite{sun2018pwc}, and our results are not as good as it in the training set, but achieve higher generalization performance in other datasets. It is difficult to obtain the ground truth of optical flow for the real dataset, so our unsupervised method has great practical application ability. DDFlow \cite{liu2019ddflow} is a multi-frame approach, and we only use two adjacent frames to achieve better results. After the UFlow \cite{jonschkowski2020matters} segments the occlusion regions, we implement fine geometric constraints of optical flow in the non-occlusion regions to achieve higher generalization performance, which indicates that better use of the essential information of optical flow in the non-occlusion regions can further improve the unsupervised performance of optical flow.

The qualitative results of our method compared with UFlow \cite{jonschkowski2020matters} on Sintel and KITTI 2015 benchmarks are shown in Fig ~\ref{vision}. It can be seen that our optical flow estimation is more uniform inside each moving objects, such as machetes, moving people, cars, grass, etc. This is because the optical flow cannot move randomly due to the proposed non-intersection and non-blocking losses inside the objects. Thus, the optical flow inside a single object is kept flexibly consistent for its motion, and the overall smoothness of optical flow for each object is ensured. At the same time, the constraints between adjacent objects also make the estimation of optical flow more accurate, such as the car motion estimation in the last two lines of Fig.~\ref{vision}. After the estimation the occlusion mask, the relative movement between adjacent objects can only be towards to cover occlusion regions. Otherwise, the relative movement of the adjacent objects can produce intersection and blocking of the optical flow. So that the relative movements of the adjacent objects are constrained, improving the optical flow estimation of adjacent objects. In addition, our visualized results are also better at detailed movements, such as pedestrian legs and butterfly movements.

\section{Conclusion}

In this paper, the motion regularity of the optical flow in the non-occlusion regions is carefully analyzed, and the geometric constraint laws of the optical flow in the non-occlusion regions are proposed. Two loss functions, non-intersection loss and non-blocking loss, are proposed based on the insight into the motion laws of optical flow in the non-occlusion regions. Their effectiveness has been proved by theoretical analysis and experiments. Optical flow is widely used in visual odometry, target tracking, dynamic segmentation, and other autonomous driving fields. The proposed method has a higher generalization performance on the real dataset, which makes the unsupervised method of optical flow in this paper have good practical application ability. Pixel-level geometric analysis and occlusion analysis are also instructive for depth estimation, visual odometry, depth completion, and scene flow estimation.

\bibliographystyle{IEEEtran}  
\bibliography{IEEEabrv,root} 

\begin{thebibliography}{10}
\providecommand{\url}[1]{#1}
\csname url@samestyle\endcsname
\providecommand{\newblock}{\relax}
\providecommand{\bibinfo}[2]{#2}
\providecommand{\BIBentrySTDinterwordspacing}{\spaceskip=0pt\relax}
\providecommand{\BIBentryALTinterwordstretchfactor}{4}
\providecommand{\BIBentryALTinterwordspacing}{\spaceskip=\fontdimen2\font plus
\BIBentryALTinterwordstretchfactor\fontdimen3\font minus
  \fontdimen4\font\relax}
\providecommand{\BIBforeignlanguage}[2]{{%
\expandafter\ifx\csname l@#1\endcsname\relax
\typeout{** WARNING: IEEEtran.bst: No hyphenation pattern has been}%
\typeout{** loaded for the language `#1'. Using the pattern for}%
\typeout{** the default language instead.}%
\else
\language=\csname l@#1\endcsname
\fi
#2}}
\providecommand{\BIBdecl}{\relax}
\BIBdecl

\bibitem{min2020voldor}
Z.~Min, Y.~Yang, and E.~Dunn, ``Voldor: Visual odometry from log-logistic dense
  optical flow residuals,'' in \emph{Proc. IEEE Conf. Comput. Vis. Pattern
  Recognit.}, 2020, pp. 4898--4909.

\bibitem{ke2018real}
R.~Ke, Z.~Li, J.~Tang, Z.~Pan, and Y.~Wang, ``Real-time traffic flow parameter
  estimation from uav video based on ensemble classifier and optical flow,''
  \emph{IEEE Transactions on Intelligent Transportation Systems}, vol.~20,
  no.~1, pp. 54--64, 2018.

\bibitem{menze2018object}
M.~Menze, C.~Heipke, and A.~Geiger, ``Object scene flow,'' \emph{ISPRS J.
  Photogram. Remote Sens. (JPRS)}, vol. 140, pp. 60--76, 2018.

\bibitem{jiang2021moving}
C.~Jiang, D.~P. Paudel, D.~Fofi, Y.~Fougerolle, and C.~Demonceaux, ``Moving
  object detection by 3d flow field analysis,'' \emph{IEEE Transactions on
  Intelligent Transportation Systems}, vol.~22, no.~4, pp. 1950--1963, 2021.

\bibitem{dosovitskiy2015flownet}
A.~Dosovitskiy, P.~Fischer, E.~Ilg, P.~Hausser, C.~Hazirbas, V.~Golkov, P.~Van
  Der~Smagt, D.~Cremers, and T.~Brox, ``Flownet: Learning optical flow with
  convolutional networks,'' in \emph{Proc. IEEE Int. Conf. Comput. Vis.}, 2015,
  pp. 2758--2766.

\bibitem{ilg2017flownet}
E.~Ilg, N.~Mayer, T.~Saikia, M.~Keuper, A.~Dosovitskiy, and T.~Brox, ``Flownet
  2.0: Evolution of optical flow estimation with deep networks,'' in
  \emph{Proc. IEEE Conf. Comput. Vis. Pattern Recognit.}, 2017, pp. 2462--2470.

\bibitem{jason2016back}
J.~Y. Jason, A.~W. Harley, and K.~G. Derpanis, ``Back to basics: Unsupervised
  learning of optical flow via brightness constancy and motion smoothness,'' in
  \emph{European Conference on Computer Vision}.\hskip 1em plus 0.5em minus
  0.4em\relax Springer, 2016, pp. 3--10.

\bibitem{ren2017unsupervised}
Z.~Ren, J.~Yan, B.~Ni, B.~Liu, X.~Yang, and H.~Zha, ``Unsupervised deep
  learning for optical flow estimation,'' in \emph{Proceedings of the AAAI
  Conference on Artificial Intelligence}, 2017.

\bibitem{wang2018occlusion}
Y.~Wang, Y.~Yang, Z.~Yang, L.~Zhao, P.~Wang, and W.~Xu, ``Occlusion aware
  unsupervised learning of optical flow,'' in \emph{Proc. IEEE Conf. Comput.
  Vis. Pattern Recognit.}, 2018, pp. 4884--4893.

\bibitem{meister2018unflow}
S.~Meister, J.~Hur, and S.~Roth, ``Unflow: Unsupervised learning of optical
  flow with a bidirectional census loss,'' in \emph{Proceedings of the AAAI
  Conference on Artificial Intelligence}, 2018.

\bibitem{alletto2018self}
S.~Alletto, D.~Abati, S.~Calderara, R.~Cucchiara, and L.~Rigazio,
  ``Self-supervised optical flow estimation by projective bootstrap,''
  \emph{IEEE Transactions on Intelligent Transportation Systems}, vol.~20,
  no.~9, pp. 3294--3302, 2018.

\bibitem{janai2018unsupervised}
J.~Janai, F.~Guney, A.~Ranjan, M.~Black, and A.~Geiger, ``Unsupervised learning
  of multi-frame optical flow with occlusions,'' in \emph{Proc. Eur. Conf.
  Comput. Vis.}, 2018, pp. 690--706.

\bibitem{liu2019ddflow}
P.~Liu, I.~King, M.~R. Lyu, and J.~Xu, ``Ddflow: Learning optical flow with
  unlabeled data distillation,'' in \emph{Proceedings of the AAAI Conference on
  Artificial Intelligence}, vol.~33, no.~01, 2019, pp. 8770--8777.

\bibitem{liu2019selflow}
P.~Liu, M.~Lyu, I.~King, and J.~Xu, ``Selflow: Self-supervised learning of
  optical flow,'' in \emph{Proc. IEEE Conf. Comput. Vis. Pattern Recognit.},
  2019, pp. 4571--4580.

\bibitem{zou2018df}
Y.~Zou, Z.~Luo, and J.-B. Huang, ``Df-net: Unsupervised joint learning of depth
  and flow using cross-task consistency,'' in \emph{Proceedings of the European
  conference on computer vision (ECCV)}, 2018, pp. 36--53.

\bibitem{ranjan2019competitive}
A.~Ranjan, V.~Jampani, L.~Balles, K.~Kim, D.~Sun, J.~Wulff, and M.~J. Black,
  ``Competitive collaboration: Joint unsupervised learning of depth, camera
  motion, optical flow and motion segmentation,'' in \emph{Proceedings of the
  IEEE/CVF Conference on Computer Vision and Pattern Recognition}, 2019, pp.
  12\,240--12\,249.

\bibitem{wang2020unsupervised}
G.~Wang, C.~Zhang, H.~Wang, J.~Wang, Y.~Wang, and X.~Wang, ``Unsupervised
  learning of depth, optical flow and pose with occlusion from 3d geometry,''
  \emph{IEEE Transactions on Intelligent Transportation Systems}, 2020.

\bibitem{jonschkowski2020matters}
R.~Jonschkowski, A.~Stone, J.~T. Barron, A.~Gordon, K.~Konolige, and
  A.~Angelova, ``What matters in unsupervised optical flow,'' in \emph{Computer
  Vision--ECCV 2020: 16th European Conference, Glasgow, UK, August 23--28,
  2020, Proceedings, Part II 16}.\hskip 1em plus 0.5em minus 0.4em\relax
  Springer, 2020, pp. 557--572.

\bibitem{geiger2012we}
A.~Geiger, P.~Lenz, and R.~Urtasun, ``Are we ready for autonomous driving? the
  kitti vision benchmark suite,'' in \emph{Proc. IEEE Conf. Comput. Vis.
  Pattern Recognit.}, 2012, pp. 3354--3361.

\bibitem{menze2015joint}
M.~Menze, C.~Heipke, and A.~Geiger, ``Joint 3d estimation of vehicles and scene
  flow.'' in \emph{ISPRS Workshop on Image Sequence Analysis (ISA)}, vol.~2,
  2015.

\bibitem{gibson1950perception}
J.~J. Gibson, ``The perception of the visual world.'' \emph{Houghton Mifflin},
  1950.

\bibitem{horn1981schunck}
B.~Horn and K.~Berthold, ``Schunck. determining optical flow,''
  \emph{Artificial Intelligence}, vol.~17, no. 1-3, pp. 185--203, 1981.

\bibitem{brox2004high}
T.~Brox, A.~Bruhn, N.~Papenberg, and J.~Weickert, ``High accuracy optical flow
  estimation based on a theory for warping,'' in \emph{Proc. Eur. Conf. Comput.
  Vis.}, 2004, pp. 25--36.

\bibitem{sun2010secrets}
D.~Sun, S.~Roth, and M.~J. Black, ``Secrets of optical flow estimation and
  their principles,'' in \emph{2010 IEEE computer society conference on
  computer vision and pattern recognition}.\hskip 1em plus 0.5em minus
  0.4em\relax IEEE, 2010, pp. 2432--2439.

\bibitem{ranjan2017optical}
A.~Ranjan and M.~J. Black, ``Optical flow estimation using a spatial pyramid
  network,'' in \emph{Proc. IEEE Conf. Comput. Vis. Pattern Recognit.}, 2017,
  pp. 4161--4170.

\bibitem{yang2019volumetric}
G.~Yang and D.~Ramanan, ``Volumetric correspondence networks for optical
  flow,'' \emph{Proc. Adv. Neural Inf. Process. Syst.}, vol.~5, p.~12, 2019.

\bibitem{butler2012naturalistic}
D.~J. Butler, J.~Wulff, G.~B. Stanley, and M.~J. Black, ``A naturalistic open
  source movie for optical flow evaluation,'' in \emph{European conference on
  computer vision}.\hskip 1em plus 0.5em minus 0.4em\relax Springer, 2012, pp.
  611--625.

\bibitem{baker2011database}
S.~Baker, D.~Scharstein, J.~Lewis, S.~Roth, M.~J. Black, and R.~Szeliski, ``A
  database and evaluation methodology for optical flow,'' \emph{International
  journal of computer vision}, vol.~92, no.~1, pp. 1--31, 2011.

\bibitem{wang2004image}
Z.~Wang, A.~C. Bovik, H.~R. Sheikh, and E.~P. Simoncelli, ``Image quality
  assessment: from error visibility to structural similarity,'' \emph{IEEE
  transactions on image processing}, vol.~13, no.~4, pp. 600--612, 2004.

\bibitem{wang2019unos}
Y.~Wang, P.~Wang, Z.~Yang, C.~Luo, Y.~Yang, and W.~Xu, ``Unos: Unified
  unsupervised optical-flow and stereo-depth estimation by watching videos,''
  in \emph{Proceedings of the IEEE/CVF Conference on Computer Vision and
  Pattern Recognition}, 2019, pp. 8071--8081.

\bibitem{zhong2019unsupervised}
Y.~Zhong, P.~Ji, J.~Wang, Y.~Dai, and H.~Li, ``Unsupervised deep epipolar flow
  for stationary or dynamic scenes,'' in \emph{Proceedings of the IEEE/CVF
  Conference on Computer Vision and Pattern Recognition}, 2019, pp.
  12\,095--12\,104.

\bibitem{liu2020flow2stereo}
P.~Liu, I.~King, M.~R. Lyu, and J.~Xu, ``Flow2stereo: Effective self-supervised
  learning of optical flow and stereo matching,'' in \emph{Proceedings of the
  IEEE/CVF Conference on Computer Vision and Pattern Recognition}, 2020, pp.
  6648--6657.

\bibitem{wang2019unsupervised}
G.~Wang, H.~Wang, Y.~Liu, and W.~Chen, ``Unsupervised learning of monocular
  depth and ego-motion using multiple masks,'' in \emph{2019 International
  Conference on Robotics and Automation (ICRA)}.\hskip 1em plus 0.5em minus
  0.4em\relax IEEE, 2019, pp. 4724--4730.

\bibitem{sun2018pwc}
D.~Sun, X.~Yang, M.-Y. Liu, and J.~Kautz, ``Pwc-net: Cnns for optical flow
  using pyramid, warping, and cost volume,'' in \emph{Proc. IEEE Conf. Comput.
  Vis. Pattern Recognit.}, 2018, pp. 8934--8943.

\bibitem{kingma2014adam}
D.~P. Kingma and J.~Ba, ``Adam: A method for stochastic optimization,''
  \emph{arXiv preprint arXiv:1412.6980}, 2014.

\end{thebibliography}

\begin{IEEEbiography}[{\includegraphics[width=1in,height=1.25in,clip,keepaspectratio]{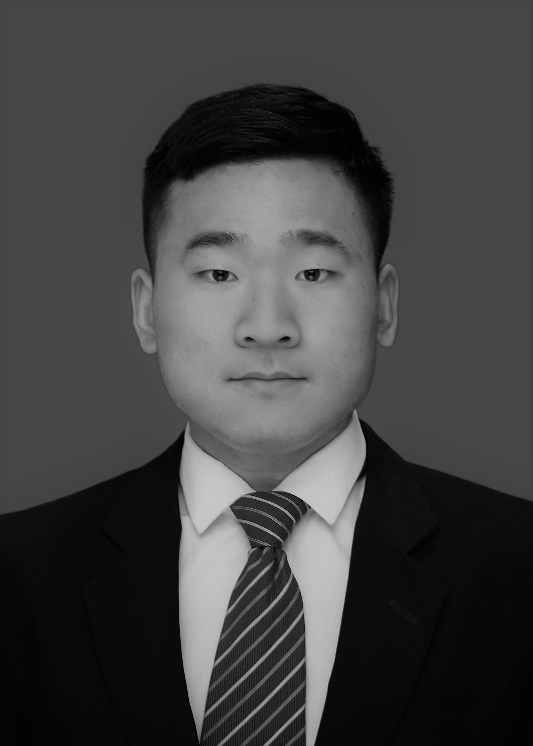}}]{Guangming Wang} (Graduate Student Member,
	IEEE) received the B.S. degree from Department of Automation from Central South University, Changsha, China, in 2018. He is currently pursuing the Ph.D. degree in Control Science and Engineering with Shanghai Jiao Tong University. His current research interests include SLAM and computer vision,  in particular, 2D optical flow estimation and 3D scene flow estimation.
\end{IEEEbiography}

\begin{IEEEbiography}[{\includegraphics[width=1in,height=1.25in,clip,keepaspectratio]{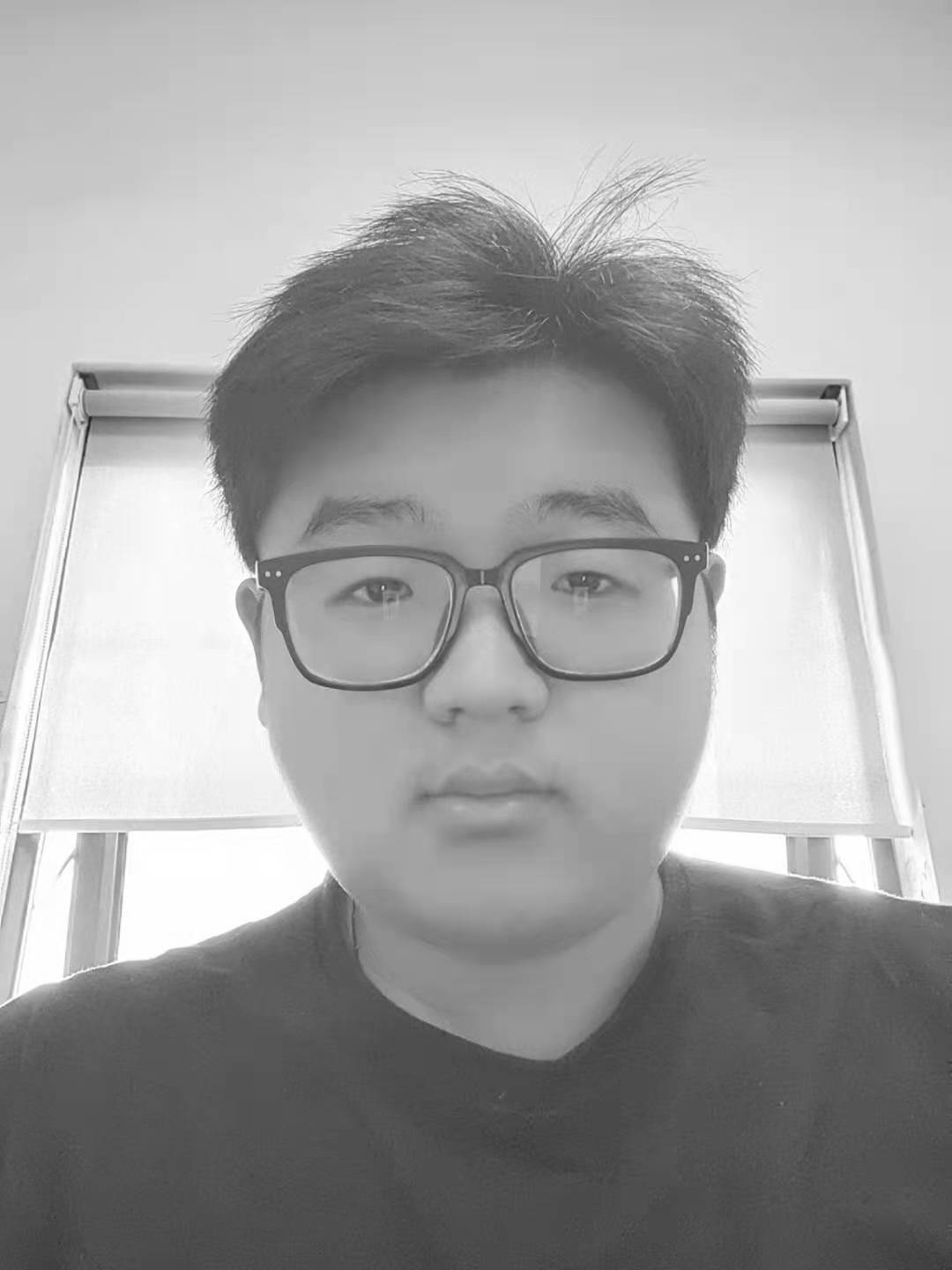}}]{Shuaiqi Ren} is currently pursuing the B.S. degree with the Department of Automation, Shanghai Jiao Tong University. His current research interests include SLAM and computer vision.
\end{IEEEbiography}

\begin{IEEEbiography}[{\includegraphics[width=1in,height=1.25in,clip,keepaspectratio]{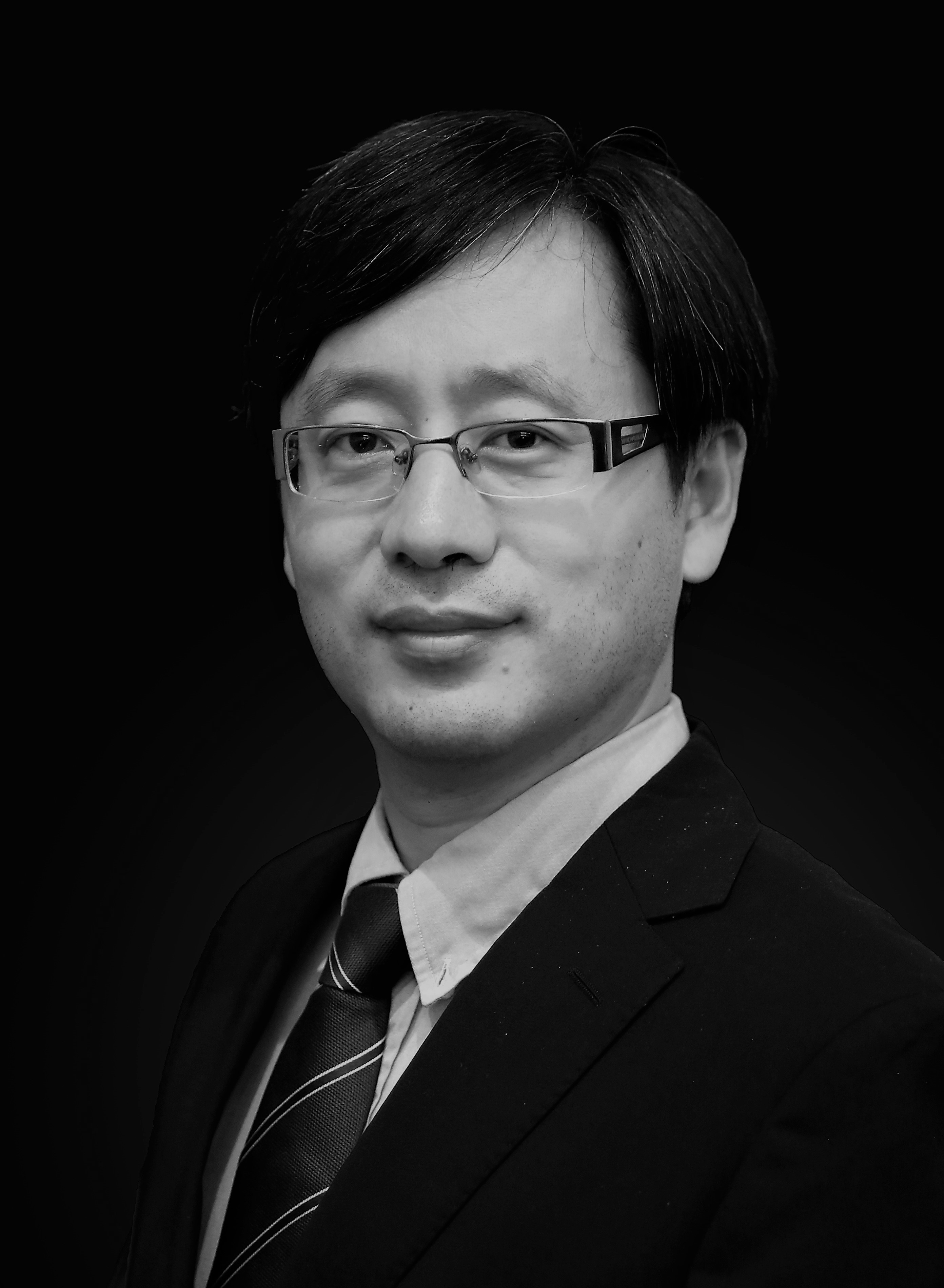}}]{Hesheng Wang}
 (Senior Member, IEEE) received the B.Eng. degree in electrical engineering from the Harbin Institute of Technology, Harbin, China, in 2002, and the M.Phil. and Ph.D. degrees in automation and computer-aided engineering from The Chinese University of Hong Kong, Hong Kong,
 in 2004 and 2007, respectively. He is currently a Professor with the Department of Automation, Shanghai
 Jiao Tong University, Shanghai, China. His current research interests include visual servoing, service robot, computer vision, and autonomous driving. He was the General Chair of the IEEE RCAR 2016, and the Program Chair of the IEEE ROBIO 2014 and IEEE/ASME AIM 2019. He has served as an Associate Editor for the IEEE TRANSACTIONS ON ROBOTICS from 2015 to 2019. He is an Associate Editor of IEEE TRANSACTIONS ON AUTOMATION SCIENCE AND ENGINEERING, IEEE ROBOTICS AND AUTOMATION LETTERS, Assembly Automation, and the International Journal of Humanoid Robotics; and a Technical Editor of the IEEE/ASME TRANSACTIONS ON MECHATRONICS. 
\end{IEEEbiography}

\end{document}